\documentclass{llncs}
\usepackage{amsmath,amssymb,bm}
\usepackage{graphicx}
\usepackage{subfigure}
\usepackage{xcolor}
\usepackage{booktabs}

\usepackage{comment}
\usepackage{bbm}
\usepackage{multirow}
\usepackage{threeparttable}
\usepackage{gensymb}
\usepackage{bbm}

\usepackage{array}
\newcolumntype{L}[1]{>{\raggedright\let\newline\\\arraybackslash\hspace{0pt}}m{#1}}
\newcolumntype{C}[1]{>{\centering\let\newline\\\arraybackslash\hspace{0pt}}m{#1}}
\newcolumntype{R}[1]{>{\raggedleft\let\newline\\\arraybackslash\hspace{0pt}}m{#1}}

\begin{document}
\title{Automated Pulmonary Nodule Detection via 3D \\ ConvNets with Online Sample Filtering and \\ Hybrid-Loss Residual Learning}

\author{Qi Dou\inst{1}, 
Hao Chen\inst{1}, 
Yueming Jin\inst{1}, 
Huangjing Lin\inst{1}, 
\\ Jing Qin\inst{2}, 
\and Pheng-Ann Heng\inst{1}} 
\institute{Dept. of Computer Science and Engineering, The Chinese University of Hong Kong
\and Center for Smart Health, School of Nursing, The Hong Kong Polytechnic University}
\maketitle

\begin{abstract}
In this paper, we propose a novel framework with 3D convolutional networks (ConvNets) for automated detection of pulmonary nodules from low-dose CT scans, which is a challenging yet crucial task for lung cancer early diagnosis and treatment.
Different from previous standard ConvNets, we try to tackle the severe hard/easy sample imbalance problem in medical datasets and explore the benefits of localized annotations to regularize the learning, and hence boost the performance of ConvNets to achieve more accurate detections.
Our proposed framework consists of two stages: 1) candidate screening, and 2) false positive reduction.
In the first stage, we establish a 3D fully convolutional network, effectively trained with an online sample filtering scheme, to sensitively and rapidly screen the nodule candidates.
In the second stage, we design a hybrid-loss residual network which harnesses the location and size information as important cues to guide the nodule recognition procedure.
Experimental results on the public large-scale LUNA16 dataset demonstrate superior performance of our proposed method compared with state-of-the-art approaches for the pulmonary nodule detection task.
\end{abstract}

\section{Introduction}
Lung cancer has been the leading cause of cancer death worldwide~\cite{setio2016validation}, and early detection of pulmonary nodules from low-dose computed tomography (CT) scans is crucial for primary lung cancer diagnosis and arrangement of early treatment.
Annual lung cancer screening for high-risk populations has already been implemented in the U.S.~\cite{national2011reduced}, acquiring enormous CT data for radiologists to analyze.
It would be difficult, if not impossible, to manually process the scans considering the huge manpower and time costs.
In this regard, accurate and cost-effective automated detection methods are highly and urgently demanded.
However, automatically detecting the pulmonary nodules is quite challenging due to the large variations in size (diameter ranging between $3\!\sim\!30$~mm), shape, density, anatomical context (e.g., solitary and lung wall/vessel/fissure attached), as well as the existence of hard mimics (i.e., tissues resembling the appearance of nodules).

Typical automated pulmonary nodule detection systems consist of two stages: 1) candidate screening, which sensitively screens candidates but gets many false positives, and 2) false positive reduction, which removes the false positive candidates and yields the final detection results.
Previously, many approaches have been proposed for this challenging detection task.
The first stage commonly relied on curvature computation, voxel clustering, intensity thresholding and morphological operations with hand-crafted features~\cite{jacobs2014automatic,murphy2009large}.
For the second stage, classifiers were utilized with low-level descriptors which were carefully and heuristically defined based on intensity, size, sphericity, texture and contextual information~\cite{jacobs2014automatic,murphy2009large,van2010comparing}.
Limited by the representation capability of low-level features,
it was difficult for these methods to deal with the large variations of nodules and distinguish them from hard mimics. 
Recent researches have proposed to rely on convolutional networks (ConvNets) for extracting high-level features to detect the pulmonary nodules~\cite{setio2016pulmonary,setio2016validation}.
Although their results are encouraging, we can still explore novel strategies such as dynamically selecting highly informative training samples and harnessing the regularization from localized annotations, to further unleash the power of ConvNets and boost the performance gains.

In this paper, we propose a novel framework leveraging 3D ConvNets in both stages to address the important while challenging task of automated pulmonary nodule detection.
Our first stage establishes a 3D fully convolutional network (FCN) to efficiently screen the candidates from volumetric CT scans.
To tackle the problem of severe hard/easy sample imbalance, we construct an online sample filtering scheme to select the highly informative training samples on-the-fly, hence to effectively train the model and enhance its discrimination capability.
In the second stage, we design a novel hybrid-loss 3D ConvNet which leverages residual learning to facilitate gradients flow and harnesses the location and size information to improve the lung nodule recognition accuracy.
Besides, our hybrid-loss residual network can also output the diameter of the detected nodule, which is clinically desirable for determining the treatment policies~\cite{setio2016validation}.
We have validated our proposed method on the large-scale LUNA16 dataset and achieved promising results outperforming those of state-of-the-art approaches.
Moreover, we have conducted extensive ablation experiments which can demonstrate the contribution of each key component in our proposed framework.

\section{Method}

\begin{figure}[t]
\centering
\includegraphics[width=0.9\textwidth]{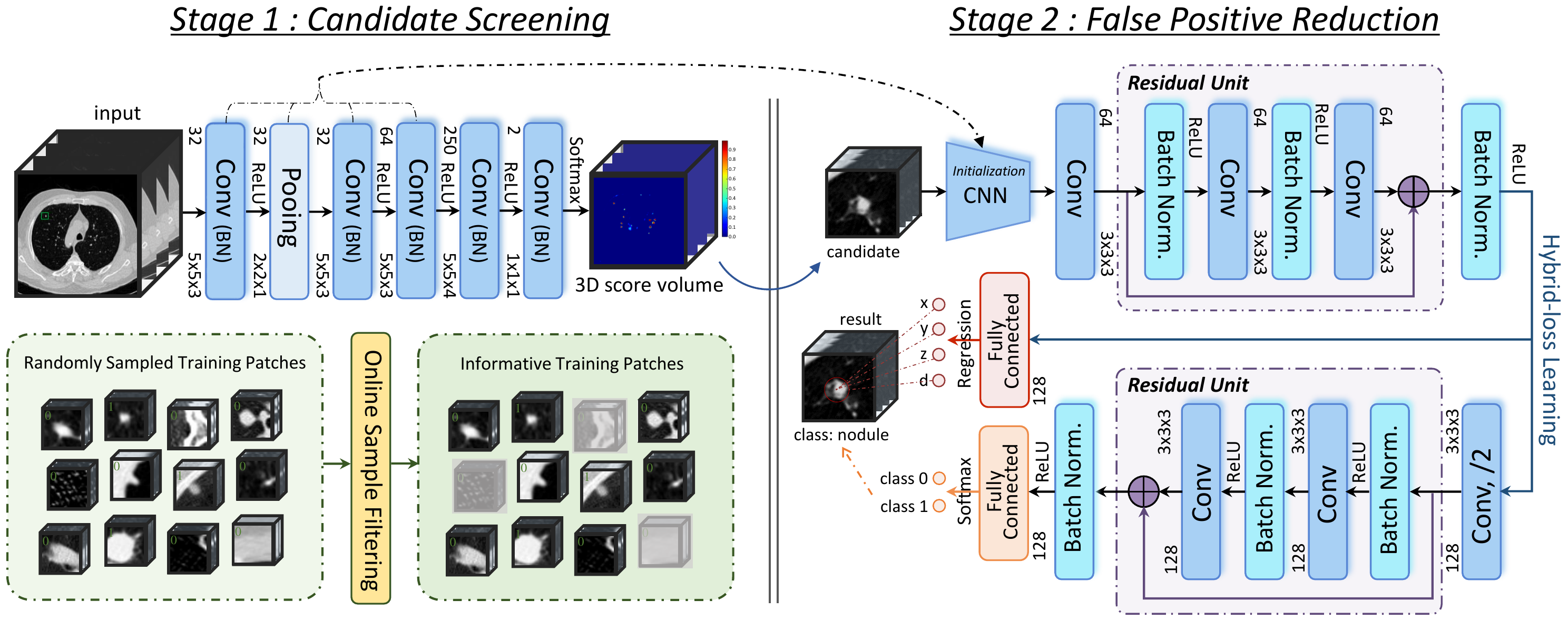}
\vspace{-2mm}
\caption{Overview of our proposed deep learning framework for automated pulmonary nodule detection, with detailed configurations of 3D ConvNets illustrated.}
\label{fig:overview}
\vspace{-4mm}
\end{figure}

Fig.~\ref{fig:overview} illustrates the proposed framework for automated detection of pulmonary nodules from low-dose 3D CT images.
We first develop a 3D FCN trained with the online sample filtering scheme to screen candidates with a high sensitivity and a fast speed.
Next, we design a hybrid-loss 3D residual network to distinguish true nodules from the candidates.

\vspace{-3mm}
\subsection{3D FCN with Online Sample Filtering for Candidate Screening}

Candidate screening throughout 3D CT scans is the initial yet crucial stage for the pulmonary nodule detection task, where both the sensitivity and efficiency are of great significance.
We establish a 3D FCN~\cite{dou2016automatic} in this stage, as it can not only leverage rich volumetric spatial information to extract high-level features for accurate candidate retrieval, but also rapidly produce the probability predictions in a volume-to-volume manner.
Specifically, we design a binary classification 3D network which consists of 5 convolutional layers and 1 max-pooling layer.
The model is trained with small 3D patches of nodules and non-nodules, and tested on the entire CT scan in a fully convolutional manner (i.e., inputing the whole image and directly getting a 3D score volume).
Then, we can retrieve candidates using the score volume with the suspicious probability of each position indicated.

However, it is challenging to train the network and achieve a high-quality score volume, given the severe imbalance between hard and easy samples.
On the one hand, an overwhelming number of background samples are quite easy to be recognized thus contribute little to optimization.
On the other hand, the number of hard samples (e.g., mimics) is considerably small, but they are difficult to be distinguished and therefore more informative for learning.
Actually, the situation of sample imbalance is a common problem in many medical detection tasks.
Previous boosting methods would consecutively train an ensemble of learners, where the misclassified samples from the former model were traced to train the next model~\cite{cirecsan2013mitosis}.
These methods repeatedly test the model on all the training data, which may complicate the training process and incur additional computations.

To tackle this problem, we propose an online sample filtering scheme to dynamically increase the proportion of hard training samples, borrowing the spirit of hard example mining originally proposed for natural object detectors~\cite{shrivastava2016training}.
Superior to previous boosting methods, the proposed scheme can select the hard samples on-the-fly during the stochastic gradient descent (SGD), and it neither interrupts the learning process nor engages additional testing computations.

Our online scheme is constructed based on the observation that hard samples usually produce higher classification loss compared with those easy ones.
In this regard, we can dynamically find the hard samples based on the loss in each forward propagation during training.
In implementation, we randomly extract the initial training samples with a large batch size.
After forward propagation of each batch, we sort the samples by their loss, and take the top $50\%$ samples on which the current model performs worst as hard samples.
Meanwhile, we randomly retain half of the remaining low-loss samples as easy samples.
These are fairly reasonable design choices, considering the balance for training patches.
Finally, we exclude those less-informative samples from the current iteration of optimization.
By performing these online modifications to SGD, our proposed scheme is intuitive yet effective to
train the ConvNet and speed up the convergence.

When the training is done, we can deploy the 3D FCN model on the whole testing CT image.
To obtain the proposal candidates, we first employ 3D non-maximum suppression on the score volume, and then retrieve those high probability positions.
Since the dimension of the score volume is reduced compared with that of the original CT image, we adopt the index-mapping mechanism in~\cite{dou2016automatic} to derive the coordinates of suspicious positions in the original image space, based on the network architecture and the coordinates in the score volume.

\subsection{Hybrid-loss 3D Residual Learning for False Positive Reduction}

Aiming to accurately distinguish true nodules from screened candidates,
we exploit a 3D ConvNet equipped with a 3D variant of residual learning technique~\cite{he2016identity} and a novel hybrid-loss objective function.
In our network, we establish a modularized 3D residual unit $\bm{x}_{\text{out}} = \bm{x}_{\text{in}} + \mathcal{F} (\bm{x}_{\text{in}}, \{W_{k}\})$, where the $\bm{x}_{\text{in}}$ and $\bm{x}_{\text{out}}$ are its input and output; the $\mathcal{F}$ is 3D residual transformation, i.e., a stack of convolutional, batch normalization and ReLU layers which are associated with the set of parameters $\{W_{k}\}$.
It has been evidenced that the residual units can boost the information flow within the network and hence benefit the optimization.

Besides the small inter-class variation between true nodules and hard mimics, another problem is that the proposal positions from FCN might deviate from the ground truth nodule centroids, due to the translation invariance inherited in the ConvNet.
This might lead to shift of true nodule locations when we center at the FCN proposal positions to crop patches for processing in the second stage.
Under this setting, leveraging localized annotations, i.e., where and how large a nodule is, would be beneficial to explicitly guide the learning to particularly focus on the targeted nodule regions.
In these regards, we design a novel hybrid-loss objective function, which considers both classification errors and localized information.
The constructed network jointly optimizes a classification branch and a regression branch, by sharing the parameters in early layers.

With a set of $N$ training pairs $\{(X^i, Y^i, G^i)\}_{i=1,...,N}$, the shared early-layer parameters $W_\text{s}$ and the classification branch weights $W_\text{cls}$ in the residual network, the classification loss is calculated as the negative log-likelihoods as follows:
\vspace{-2mm}
\begin{equation}
\vspace{-1mm}
\mathcal{L}_{\text{cls}} = - \frac{1}{N}\sum_i\log p(Y^i|X^i,W_\text{s},W_\text{cls}).
\end{equation}
\vspace{-3mm}
\\
For the regression branch, considering that we are targeting 3D objects, our localization ground truth $G^i=(G^i_x,G^i_y,G^i_z,G^i_d)$ is represented by four parameters, with $(G^i_x, G^i_y, G^i_z)$ and $G^i_d$ respectively being the centroid and diameter of the nodule.
Denoting the 3D FCN proposal position by $P^i=(P^i_x,P^i_y,P^i_z)$, and the second stage cropped patch size by $S=(S_x, S_y, S_z)$, we define the continuous-valued regression target $T^i=(T_x^i, T_y^i, T_z^i, T_d^i)$ as:

\begin{equation}{\small}
\resizebox{0.93\textwidth}{!}{$
T^i_x \! = \! \frac{2(G^i_x-P^i_x)}{S_x},  T^i_y \! = \! \frac{2(G^i_y-P^i_y)}{S_y},   T^i_z \! = \! \frac{2(G^i_z-P^i_z)}{S_z}, 
T^i_d \! =  \! \log (\frac{G^i_d}{\sqrt{S_x^2+S_y^2+S_z^2}}),
$}
\vspace{-1mm}
\end{equation}
where $T^i$ specifies a scale-invariant translation and log-space size shift relative to the cropped patch size $S$.
Considering the candidate proposal $P^i$ is near to the ground truth nodule centroid, we divide their relative distance with half patch size for normalization.
Denoting the output of the regression branch by $\hat{T^i} \! = \! f(X^i,W_s,W_\text{reg})$,
the loss from location information of training sample $i$ is:
\vspace{-1mm}
\begin{equation}
\mathcal{L}^i_{\text{loc}} ~ = \sum_{\gamma \in \{x,y,z,d\}}  \mathbbm{1}(Y^i=1) \cdot \text{dist} (T^i_\gamma - \hat{T}^i_\gamma),
\vspace{-1mm}
\end{equation}
where the function $\text{dist}(a) = 0.5a^2$ if $|a|<1$, otherwise $|a|-0.5$, which is a robust $L_1$ loss and less sensitive to outliers than the $L_2$ loss~\cite{Girshick_2015_ICCV}.
The $\mathbbm{1}(Y^i=1)$ is the indicator function, with which we only consider the localization loss for positive samples, and ignore those non-nodule samples without size notion.
Overall, our hybrid-loss objective function is formulated as follows:
\vspace{-1mm}
\begin{equation}
\mathcal{L} = \mathcal{L}_{\text{cls}} + \lambda \frac{1}{N_{\text{reg}}} \sum_{i} \mathcal{L}^i_{\text{loc}} + \beta (||W_\text{s}||^2_2 + ||W_\text{cls}||^2_2 + ||W_\text{reg}||^2_2).
\vspace{-1mm}
\end{equation}
The first term is nodule classification loss.
The second term is localization loss, where the $N_{\text{reg}}$ represents the number of samples considered in the regularization.
The third term is weight decay of the shared, classification and regression parameters.
The $\lambda$ and $\beta$ are balancing weights.
Note that the proposed hybrid-loss residual network can not only improve the detection accuracy by introducing guidance from location information, but also output the size of detected nodule, which is clinically desirable from automated methods with diagnosis significance.

\vspace{-3mm}
\section{Experimental Results}

We validated the proposed framework on the large-scale public challenge dataset of LUNA16~\cite{setio2016validation}, which contains $888$ low-dose CT scans with the location centroids and diameters of the pulmonary nodules annotated.
\\
\\
\textbf{Implementation Details.}
We conducted augmentations for positive samples, including random translation within radius region of the nodule, flipping, random scaling between $[-0.9,+1.1]$, and random rotation of $[90\degree, 180\degree, 270\degree]$ in the transverse plane.
We set a relatively small training patch size ($30\!\times\!30\!\times\!10$) in the first stage for fast screening 
and  the second stage employed a larger size ($60\!\times\!60\!\times\!24$) to include richer contextual information for more accurate detection.
The 3D FCN model was randomly initialized from a Gaussian distribution $\mathcal{N}(0,0.01)$, and the learning rate was initialized as $0.001$.
The score volume threshold for candidate screening was set as $0.85$, which was determined via grid search on a validation set.
When training the hybrid-loss residual network, the first three convolutional layers were initialized from the FCN model and the rest parameters of deeper layers were randomly initialized as in~\cite{he2016identity}.
The convolutions in the residual units used padding to maintain the dimension of feature volumes.
The $\lambda$ and $\beta$ were set as $0.5$ and 1e-4, respectively.
We implemented our method in the Lasagne framework with Theano backend using an Nvidia Titan X GPU.
\begin{table}[t]
\caption{Comparison with different nodule detection methods on LUNA16 dataset~\cite{setio2016validation}.}
\vspace{-2mm}
\centering
\resizebox{\textwidth}{!}{ %
\begin{tabular}{l|C{1cm}C{1cm}C{1cm}C{1cm}C{1cm}C{1cm}C{1cm}|c}
    \toprule[1pt]
    Teams                          & 0.125 & 0.25 & 0.5 & 1 & 2 & 4 & 8 & ~CPM Score \\
    \midrule
    DIAG\_ConvNet~     & \textbf{0.692} & \textbf{0.771} & 0.809 & 0.863 & 0.895 & 0.914 & 0.923 & 0.838 \\
    ZENT~            & 0.661 & 0.724 & 0.779 & 0.831 & 0.872 & 0.892 & 0.915 & 0.811 \\
    Aidence~         & 0.601 & 0.712 & 0.783 & 0.845 & 0.885 & 0.908 & 0.917 & 0.807 \\
    MOT\_M5Lv1~       & 0.597 & 0.670 & 0.718 & 0.759 & 0.788 & 0.816 & 0.843 & 0.742 \\
    VisiaCTLung~      & 0.577 & 0.644 & 0.697 & 0.739 & 0.769 & 0.788 & 0.793 & 0.715 \\
    Etrocad~         & 0.250 & 0.522 & 0.651 & 0.752 & 0.811 & 0.856 & 0.887 & 0.676 \\
    \textbf{Our Method}  & 0.659 & 0.745 & \textbf{0.819} & \textbf{0.865} & \textbf{0.906} & \textbf{0.933} & \textbf{0.946} & \textbf{0.839} \\
    \bottomrule[1pt]
\end{tabular}}
\label{tab:comp}
\vspace{-5mm}
\end{table}
\\
\textbf{Detection Results and Comparison with Other Methods.}
The LUNA16 dataset was randomly divided into ten subsets for ten-fold cross validation.
The evaluation metrics are sensitivity and average number of false positives per scan (FPs/scan).
A detection is counted as true positive if it locates within the radius of a nodule centroid.
The challenge also computes a CPM score (i.e., the average sensitivity at 7 predefined FPs/scan rates: 1/8, 1/4, 1/2, 1, 2, 4, 8) to assess the performance of each algorithm.
Table~\ref{tab:comp} reports the detection performance of our method and that of other approaches in the challenge.
In fact, all participates employed deep learning approaches, and we refer readers to~\cite{setio2016validation}, a comprehensive summary of LUNA16, to learn more details of other methods.
It is observed in Table~\ref{tab:comp} that our method achieves a CMP of $0.839$, outperforming other state-of-the-art methods.
In addition, we achieve the best results at five of the seven predefined FPs/scan rates, corroborating the effectiveness of our novel learning strategies with online sample filtering and hybrid-loss residual network. 
It is reported that, in clinical practice, the FPs/scan rates between $1$ and $4$ are mostly concerned~\cite{van2010comparing}.
Our method yields a sensitivity of $90.6\%$ at $2$ FPs/scan, highlighting its promising potential to be exploited in clinical practice.
\\
\begin{figure}[t]
\centering
	\begin{minipage}[t]{0.49\textwidth}
	\includegraphics[width=1.0\linewidth]{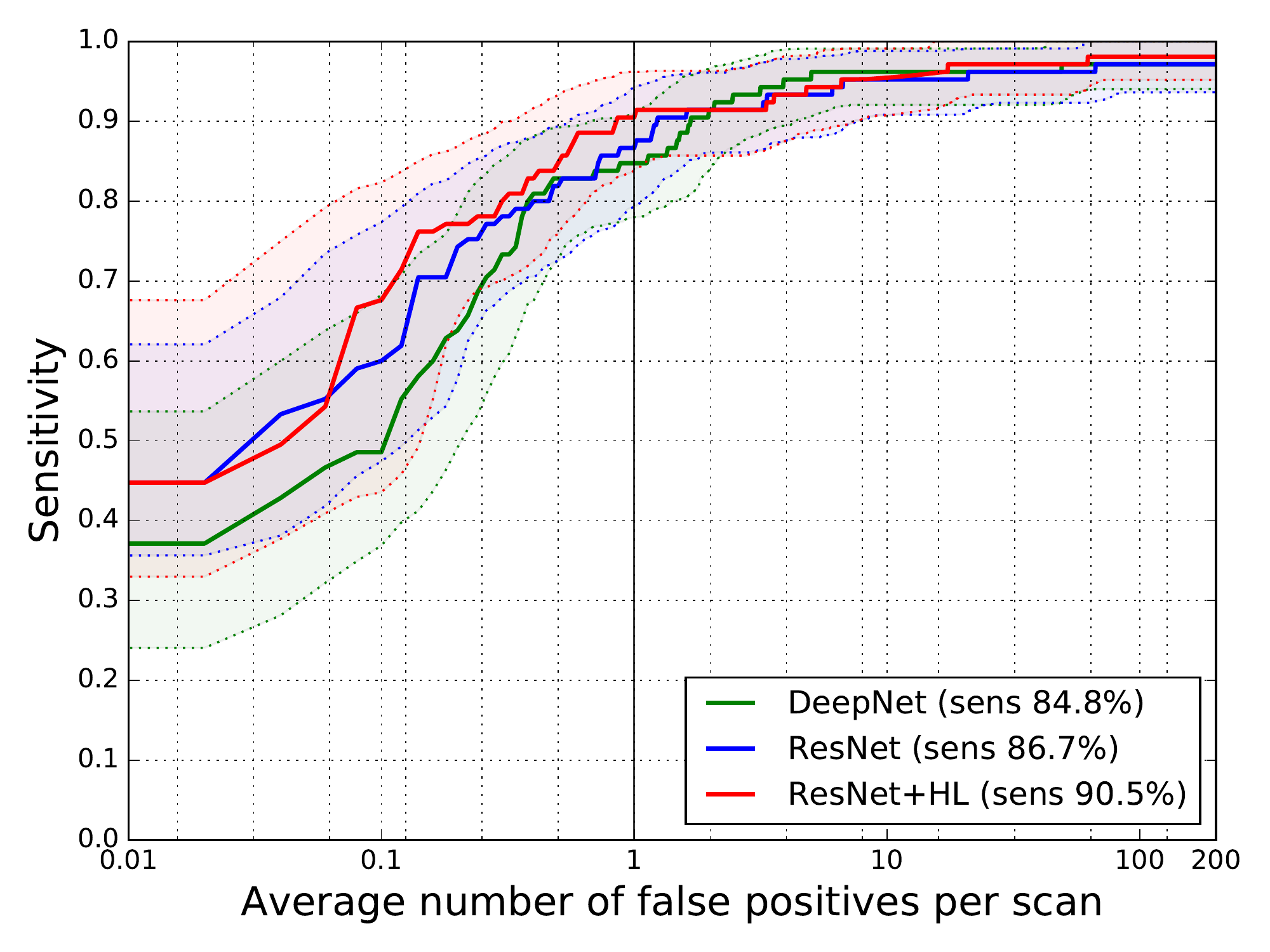}
	\end{minipage}
	\begin{minipage}[t]{0.49\textwidth}
	\includegraphics[scale=0.196]{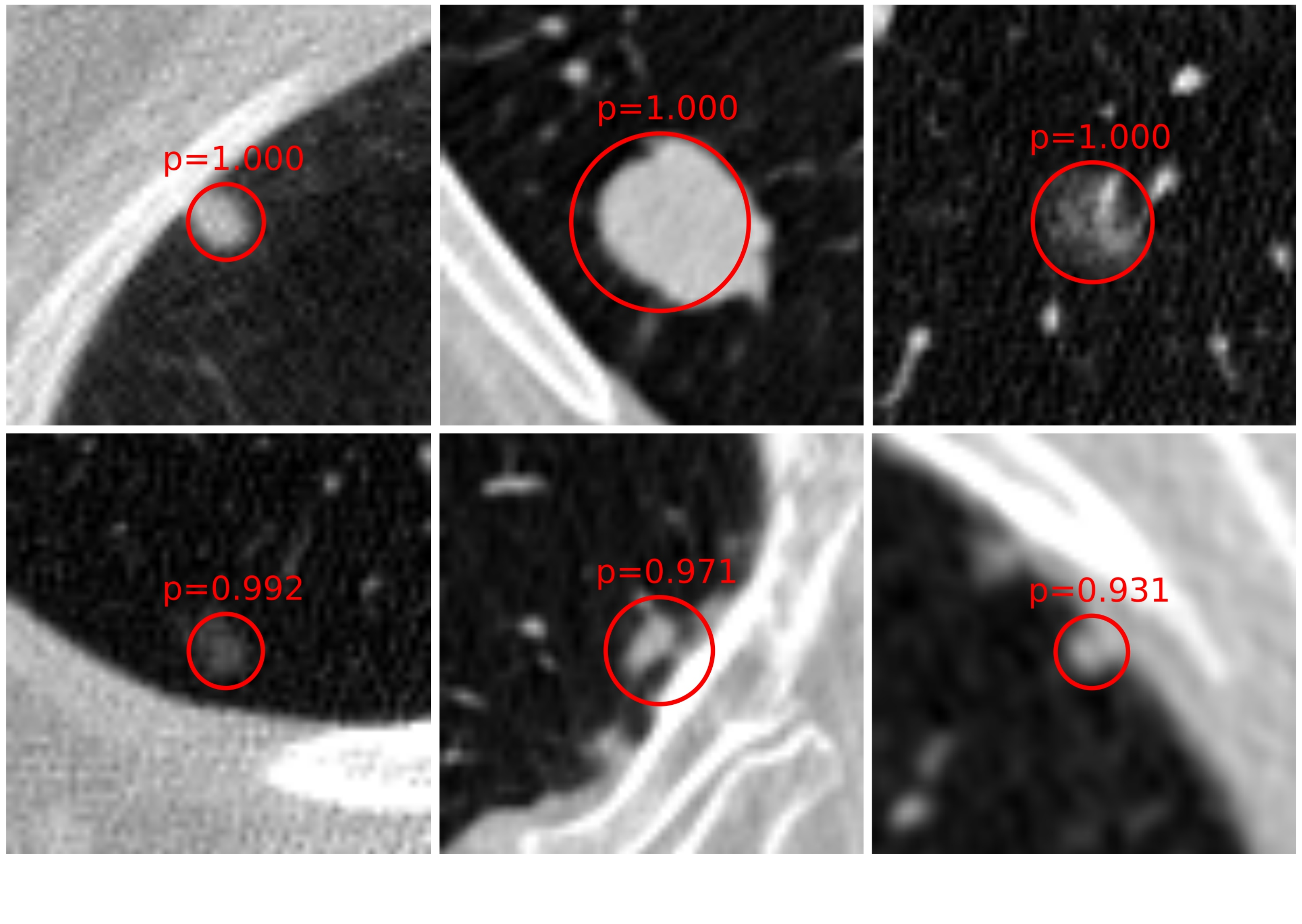}
	\end{minipage}
\vspace{-2mm}
\caption{Left: Comparison of FROC curves using different network configurations, with shaded areas presenting the 95\% confidence interval. Right: Examples of detection results from our method with the prediction probability and diameter indicated in red.}
\label{fig:comp}
\end{figure}
\\
\textbf{Ablation Studies.}
To further investigate the contribution of each key component in our framework, we performed extensive ablation experiments, using nine subsets for training and one subset for testing.
We first assess the capability of screening nodule candidates using the 3D FCN trained to convergence with and without online sample filtering (OSF) scheme.
The results are listed in the first two columns of Table~\ref{tab:result}.
The OSF training significantly improves the candidate screening performance by increasing the sensitivity from 94.3\% to 97.1\% and reducing the FPs/scan from 286.2 to 219.1.
These improvements demonstrate that selecting the high-loss samples (hard samples) with OSF scheme can greatly enhance the model's discrimination capability.
To further evaluate the effectiveness of the residual learning technique and the hybrid-loss objective equipped in our ConvNet for false positive reduction, we implemented three networks, i.e., plain deep network (DeepNet), residual network (ResNet), and the proposed hybrid-loss residual network (ResNet+HL), according to the architecture shown in Fig.~\ref{fig:overview}.
Their results are presented in the last three columns of Table~\ref{tab:result}.
With $1.0$ FPs/scan, the three networks achieve detection sensitivities of $84.8\%$, $86.7\%$, and $90.5\%$, respectively, demonstrating that while residual learning technique can improve the performance of traditional ConvNets by facilitating gradients flow in optimization, the proposed hybrid-loss objective function can further boost the detection performance by additionally supervising the training with location and size information.
Fig.~2 (left) presents the three networks' free-response receiver operating characteristic (FROC) curves for more comprehensive comparison at a wider range of false positive rates.
\begin{table}[t]
\caption{Evaluation of each component in our pulmonary nodule detection framework.}
\vspace{-2mm}
\centering
\resizebox{\textwidth}{!}{ %
\begin{tabular}{l|C{2cm}|C{2cm}| C{2cm}|C{2cm}|C{2cm}}
\toprule
Stages & \multicolumn{2}{c|}{Candidate Screening} & \multicolumn{3}{c}{False Positive Reduction}  \\
\hline
Methods      & FCN     & FCN+OSF         & DeepNet & ResNet & ResNet+HL\\
\hline
Sensitivity  & 94.3\%  & \textbf{97.1\%} & 84.8\%  & 86.7\% &  \textbf{90.5\%} \\
FPs/scan     & 286.2   & \textbf{219.1}  & 1.0     & 1.0    &  1.0 \\
\bottomrule
\end{tabular}
\label{tab:result}}
\vspace{-6mm}
\end{table}
It is observed that the ResNet+HL continually obtains the best performance among the three configurations.
The Fig.~\ref{fig:comp} (right) depicts typical examples of final detection results with the classification probability and regressed diameter indicated.
We can find that our model can recognize the various nodules with a high probability, as well as reliably output the size of the detected nodules.
Last but not least, our proposed detection framework is quite efficient
taking less than one minute for one subject, which enables our method to be competent for large-scale data processing, such as the annual lung cancer screening program launched for high-risk populations.

\section{Conclusion}
This paper proposes a novel framework with 3D ConvNets to automatically detect pulmonary nodules in low-dose 3D CT scans.
For candidate screening, we establish a 3D FCN trained with an online sample filtering scheme which can effectively tackle the hard/easy sample imbalance problem.
For false positive reduction, we design a hybrid-loss residual ConvNet, which harnesses the localized annotations to improve nodule classification accuracy.
Extensive experiments validate the efficacy of our detection framework.
Note that as tackling sample imbalance and harnessing localized information are both common issues when employing ConvNets in medical image computing tasks, our method is extensible to many other applications and could inspire more intelligent learning strategies.
\\
\\
\textbf{Acknowledgments.} This work was supported by the following grants from the Research Grants Council (Project no. CUHK412513) and the Innovation and Technology Fund (Project no. ITS/041/16) of Hong Kong.

\bibliographystyle{splncs03}
\bibliography{refs}

\end{document}